\documentclass[conference]{IEEEtran}
\IEEEoverridecommandlockouts
\usepackage{cite}
\usepackage{amsmath,amssymb,amsfonts}
\usepackage{algorithmic}
\usepackage{graphicx}
\usepackage{textcomp}
\usepackage{xcolor}
\usepackage{amsmath}
\usepackage{url}
\renewcommand{\arraystretch}{}
\def\BibTeX{{\rm B\kern-.05em{\sc i\kern-.025em b}\kern-.08em
    T\kern-.1667em\lower.7ex\hbox{E}\kern-.125emX}}
\begin{document}

\title{Weakly-supervised Autism Severity Assessment in Long Videos\\
\thanks{* Corresponding Authors \\
COFUND BoostUrCAreer program received funding from the European Union’s Horizon 2020 under Marie Curie grant agreement No 847581. This work is also supported by the
French government, through the ACTIVIS project managed by the National Research Agency (ANR) with the reference number ANR-19-CE19-0004.}
}

\author{\IEEEauthorblockN{Abid Ali}
\IEEEauthorblockA{\textit{STARS Team, INRIA, Sophia Antipolis} \\
Valbonne, France \\
abid.ali@inria.fr}
\and
\IEEEauthorblockN{Mahmoud Ali}
\IEEEauthorblockA{\textit{STARS Team, INRIA, Sophia Antipolis} \\
Valbonne, France \\
mahmoud.ali@inria.fr}
\and
\IEEEauthorblockN{Camilla Barbini}
\IEEEauthorblockA{\textit{CoBTek, Université Côte d’Azur} \\
Nice, France \\
camilla.barbini@hpu.lenval.com}
\and
\IEEEauthorblockN{Séverine Dubuisson}
\IEEEauthorblockA{\textit{LIS} \\
Marseille, France \\
severine.dubuisson@lis-lab.fr}
\and
\IEEEauthorblockN{Jean-Marc Odobez}
\IEEEauthorblockA{\textit{Idiap Research Institute} \\
% \textit{Weakly Supervised Autism Severity Assessment in Long Videos}\\
Martigny, Switzerland \\
jean-marc.odobez@idiap.ch}
\and
\IEEEauthorblockN{Francois Bremond*}
\IEEEauthorblockA{\textit{STARS Team, INRIA, Sophia Antipolis} \\
Valbonne, France \\
francois.bremond@inria.fr}
\and
\IEEEauthorblockN{Susanne THÜMMLER*}
\hspace{16cm}\IEEEauthorblockA{\textit{CoBTek, Université Côte d’Azur} \\
Nice, France \\
susanne.thummler@hpu.lenval.com}
}

\maketitle

\begin{abstract}
Autism Spectrum Disorder (ASD) is a diverse collection of neurobiological conditions marked by challenges in social communication and reciprocal interactions, as well as repetitive and stereotypical behaviors. Atypical behavior patterns in a long, untrimmed video can serve as biomarkers for children with ASD. In this paper, we propose a video-based weakly-supervised method that takes spatio-temporal features of long videos to learn typical and atypical behaviors for autism detection. On top of that, we propose a shallow TCN-MLP network, which is designed to further categorize the severity score. We evaluate our method on actual evaluation videos of children with autism collected and annotated (for severity score) by clinical professionals. Experimental results demonstrate the effectiveness of behaviors biomarkers that could help clinicians in autism spectrum analysis.
\end{abstract}


\section{Introduction}
Autism Spectrum Disorder (ASD) is a diverse collection of neurobiological conditions marked by challenges in social communication and reciprocal interactions, as well as repetitive and stereotypical behaviors. ASD typically manifests in early childhood and significantly impacts the lives of affected children and their families, with no established cure currently available. Although ASD is linked to a variety of factors, including genetics, biology, and environmental influences, the exact causes remain unidentified in many patients \cite{o2008autism}. Additionally, the incidence of ASD is increasing. According to the World Health Organization (WHO), 1 in 100 children has ASD \cite{WHO}. This figure is an average derived from multiple studies, which report a wide range of prevalence rates. According to data from the Autism and Developmental Disabilities Monitoring (ADDM) network in 2016, the current prevalence of autism spectrum disorder is one in every 54 children \cite{knopf2020autism}. Furthermore, the rate of ASD in middle- and low-income countries remains undetermined.

In a clinical setting, autism is identified through an interactive session where a skilled healthcare professional evaluates specific behavioral characteristics using both verbal and non-verbal tasks. The literature generally agrees that early detection, coupled with ongoing intervention, is crucial to optimize therapeutic outcomes. Therefore, taking advantage of brain neuroplasticity during early childhood, prompt diagnosis of ASD, and suggesting comprehensive behavioral interventions can lead to improved long-term results. Nonetheless, diagnosing ASD remains a complex task. Key factors involve specialized knowledge and specific diagnostic instruments that rely on interpreting child behavior, conducting parent interviews, long-term monitoring and symptom examination, and manual analysis. These assessments are time-consuming and clinically require arduous processes. Moreover, human evaluations can be subjective and vary widely. Effective treatment necessitates prompt diagnosis, yet accurate evaluations are typically not made until age 5, which is considered late for intervention \cite{hashemi2012computer}. There is a need for a more appropriate and accessible initial diagnosis to enhance the accuracy of ASD detection.
% Communication and language assessments are crucial elements in the diagnosis process [17]. However, about 40 percent of children with autism do not use verbal communication, adding another layer of complexity to the diagnostic process for this particular group [4].An early diagnosis could guarantee early intervention and access to tailored therapies for efficient management.
 % Initial research indicated that irregularities in social interactions, communication, stereotypical, and the display of repetitive behaviors could be the main signs of ASD. However, some symptoms related to communication and behavior, often associated with autism, are not unique to ASD, complicating the early diagnosis. \cite{lewis1998repetitive}. Effective treatment necessitates prompt diagnosis, yet accurate evaluations are typically not made until age 5, which is considered late for intervention \cite{hashemi2012computer}. There is a need for a more appropriate and accessible initial diagnosis to enhance the accuracy of ASD detection.

Throughout the years, researchers have proposed several methods for ASD detection \cite{al2021gait, boutrus2019increased, chen2019attention, ali2022video, li2022appearance, negin2021vision}. Many of these methods focused mainly on a single module such as either repetitive gesture analysis (skeleton-based or appearance-based) or facial or eye-gaze patterns. However, a single module do not provide detailed insight into autistic behavior traits such as emotion exchanges, social-communication difficulties, atomic stereotypes, unusual or unbalanced movements, etc., which together form a crucial part of the diagnosis \cite{healthdirect2021} process. Recent studies indicate that children with autism often display unique biomarkers of gestures, facial and emotional expressions, and behavioral activities. Utilizing these biomarkers can aid in identifying a distinct distribution of features, thereby enhancing the evaluation of autism.

Distinctive behavior biomarkers in children with autism may encompass \textbf{stimming or repetitive movements} such as flapping, rocking, specific \textbf{atomic hand gestures} such as playing with hair, mouth and nose, etc., and \textbf{limited gestures} coupled with challenges in interpreting others’ gestures. They may also exhibit \textbf{unusual or unbalanced movements} and \textbf{impaired motor coordination}, leading to difficulties in fine motor skills such as grasping and holding objects, and gross motor skills like jumping and balancing.

Assessing Autism Spectrum Disorder (ASD) by collectively evaluating all the above-mentioned behavioral biomarkers presents a significant challenge. The scarcity of available data in existing literature compounds this difficulty. Current public datasets primarily concentrate on specific aspects such as repetitive movements, as seen in SSBD\cite{rajagopalan2013self} and ESSBD\cite{negin2021vision}, or on facial expressions and eye-gaze patterns, as in the case of MMBD\cite{rehg2013decoding}. Additionally, certain datasets such as De-Enigma\cite{marinoiu20183d} are not publicly accessible.

In this paper, we propose a video-based weakly-supervised method that leverages spatio-temporal features of a long video to learn typical and atypical behavior patterns for autism detection. The resulting weakly-supervised network is further exploited to train a shallow regression model in a supervised manner to infer different severity levels according to the Autism Diagnostic Observation Schedule (ADOS) protocol. 

We evaluate our method on actual evaluation videos of children with autism collected and annotated by clinical professionals. Experimental results demonstrate the effectiveness of spatio-temporal behavior patterns in accurately identifying autistic children. This could greatly influence the early detection and treatment of ASD by offering a dependable, non-disruptive, and effective means for autism categorization. Furthermore, the focus on actions simplifies the evaluation of children with restricted verbal communication. To sum-up, the main contributions are as follows:
\begin{itemize}
    \item We propose a weakly-supervised network to learn discriminative markers in untrimmed videos related to typical and atypical behaviors.
    \item Our severity score regressor module can automatically regress the autism severity score according to ADOS.
    \item We evaluate our method on real-world autism assessment videos.
\end{itemize}

\section{Related Work}
Current studies have investigated diverse methods for autism evaluation, with a significant focus on techniques based on facial expressions, eye gaze patterns, and gestures. 

\subsection{Action Detection} 

Temporal Action Localization (TAL) is a fundamental task in video understanding.  In terms of supervised methods, \cite{farha2019ms} proposed a multi-stage architecture for temporal action segmentation. The first stage generates an initial prediction which is refined by the next stages. PDAN \cite{dai2021pdan} introduces a Dilated Attention Layer (DAL) for allocating attention weights to local frames and constructs a pyramid of DALs with different dilation rates to capture both short-term and long-term temporal relations. In this work, we experiment with PDAN \cite{dai2021pdan} and MS-TCN \cite{farha2019ms} for SOTA comparison on supervised methods. However, such a fully supervised setting suffers from limitations like expensive frame-level labeling and subjective, prone to manual errors. 

On the other hand, Weakly Supervised Temporal Action Localization (WTAL) methods have been developed. WTAL involves classifying and localizing all action instances in untrimmed videos under the supervision of only video-level category labels. \cite{joo2023clip} utilizes ViT-encoded visual features from CLIP \cite{radford2021learning} to extract discriminative representation and models temporal dependencies using Temporal Self-Attention (TSA). The OE-CTST \cite{majhi2024oe} enhances the CLIP-TSA \cite{joo2023clip} by introducing an anomaly-aware temporal position encoding and a cross-temporal scale transformer. Our idea is borrowed from OE-CTST \cite{majhi2024oe} for autistic behavioral coding. 

The majority of Temporal Action Localization (TAL) techniques frequently take advantage of large-scale Foundation Models (FMs) to extract high-dimensional features. In this work, we experiment with the DinoV2 \cite{oquab2023dinov2} and the VideoMAE-v2\cite{wang2023videomae} features to understand atypical ASD behaviors in untrimmed videos. 

\subsection{Facial and Eye-Gaze Based}
Physical appearance is a distinguishable characteristic of
autism. In \cite{tariq2019detecting}, developmental setbacks can be discerned from physical appearances in home-recorded videos. Asymmetry in facial appearance is studied in \cite{rehg2013decoding}. The research indicates that people with a history of ASD often exhibit more asymmetric features. 
% Other studies also corroborate that children with ASD have greater facial asymmetry \cite{}. 
The pattern of eye-gaze is also a significant indicator of autism, as children with ASD tend to exhibit less attention compared to typically developed children \cite{riby2009looking}. Their facial expressions and direction of gaze do not interact with their environment. This pattern of reduced eye gaze is consistently observed in all age groups and cultures \cite{ma2021atypical}. The cumulative stack histogram, as suggested in \cite{li2020classifying}, identifies these irregularities in the trajectory of eye movement. AttentionGazeNet \cite{li2022appearance} creates a mapping of screen coordinates from 3D gaze vectors. Experimental results suggest that gaze vectors are more scattered in children with ASD. 

However, recognizing ASD from facial and eye gaze analysis is limited to only a few cues of autism, neglecting other atypical behaviors such as uncontrolled or limited body movements, impaired motor coordination and repetitive behaviors, etc.  

\begin{figure*}[t]
\centering
\includegraphics[width=0.97\linewidth]{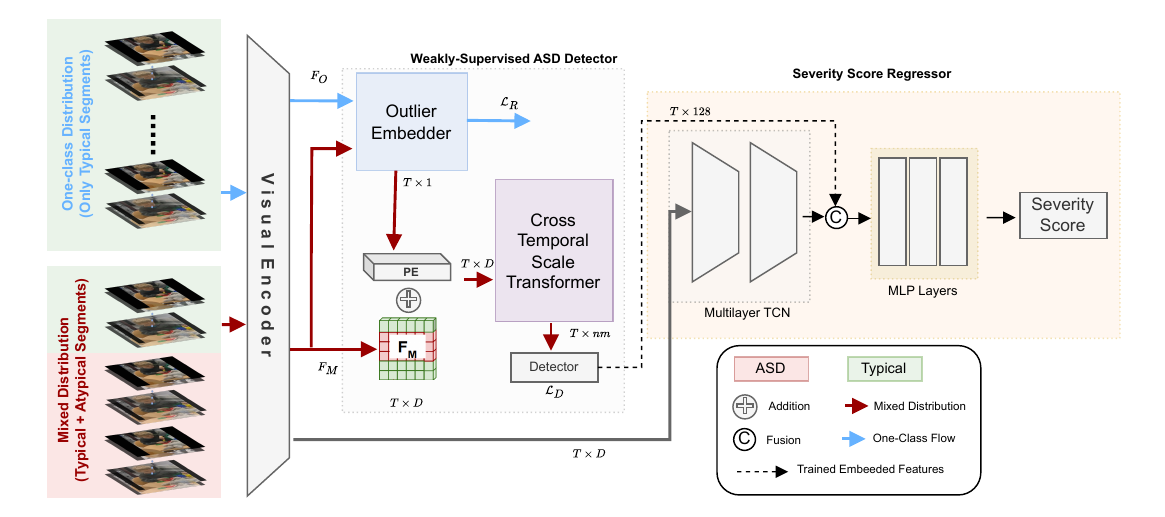}
\caption{The network comprises three major stages \textit{i.e.} (A) Visual Encoder, (B) Weakly-supervised ASD Detector to detect typical and atypical behaviors, (C) Severity Score Regressor to further regress the final severity score. Here, $F_O$ = feature map of one-class, $F_M$ = feature map of mixed distribution, $T$ = 32 temporal segments, $D$ = 1408 features, $128$ is the feature vector from detector final layer. $nm$ is the $m$ video features obtained from $n$-levels of CTST module.}
\label{fig:model}
\end{figure*}

\subsection{Gesture-Based}
The study \cite{anzulewicz2016toward} reveals a notable difference in hand gesture patterns between children with Autism Spectrum Disorder and those who are typically developing. When these children engage in games on a smart tablet, those with ASD tend to apply more force and pressure in their gestures, and also utilize a larger average area. Another study \cite{zunino2018video}, proposes that differences in gesture patterns when performing actions may also be apparent from the very beginning, incorporating information about intention. Therefore, the intended gestures can serve as a diagnostic tool for children with ASD. These studies underscore the potential to use motor functions in the analysis of ASD.

In the study \cite{al2021gait}, features crafted from skeletal data are utilized to categorize children with ASD. The attention-focused ASD screening technique in \cite{chen2019attention} leverages various modalities to incorporate complementary multimodal information into a common space. 

Another line of research is centered on identifying atypical actions from videos. The approach known as Bag-of-visual-words [30] interprets image grids as visual words to identify pertinent feature descriptors. \cite{ali2022video} employed a two-stream architecture to classify repetitive autistic actions. In \cite{tian2019video}, a temporal pyramid network is employed to generate layers of feature maps from long-duration videos. A distinct discriminator for repetitive behavior is utilized to enhance the training process by differentiating samples that exhibit unusual actions. 

Though extensive research has been done on skeleton and appearance-related approaches, these approaches are limited to short gestures of a few seconds such as jumping, flapping, and/or rocking, etc. They mostly use end-to-end deep learning methods and do not incorporate attention to the underlying mechanisms of atypical behaviors in children with ASD. Thus, in this study, we delve into the atypical behavioral patterns present in long videos and assimilate them into the learning process to amplify the representation of discriminative markers.

\section{Method}
The architecture we propose is comprised of three distinct stages. In the initial stage, we extract features at the video-level from each untrimmed video. Subsequently, we employ a weakly-supervised method to classify autistic and typical children. Ultimately, we train a shallow architecture to derive the final severity score for each individual.

\subsection{Visual Encoder}
The primary goal of the visual encoder is to derive spatio-temporal features from long, untrimmed videos. Initially, the input video $V$ is split into $T$ non-overlapping consecutive temporal segments, each containing a series of $16$ successive frames. For each segment, we utilize a VideoMAE-v2 \cite{wang2023videomae} architecture to generate a feature map of dimension $1 \times D$. Each segment-level feature can be interpreted as a temporal token, and for a given $V$ with $T$ segments, the visual encoder produces a video feature map of dimension $T \times D$. During the training phase, the visual encoder produces two batches of video feature maps i.e., one from typical and the other from mixed distribution "mixed includes both typical and atypical segments", denoted as $F_O$ and $F_M$ respectively, which are then processed by OE and CTST modules of the weakly-supervised method \cite{majhi2024oe}. 
% We experiment with Dinov2 \cite{dino} and VideoMAE-v2 \cite{vid} for feature extraction. 

\subsection{Weakly-Supervised Autism Detection}
We borrowed OE-CTST \cite{majhi2024oe}, a WTAL anomaly detection architecture, to learn the atypical and typical behavioral patterns of children with and without ASD. The architecture consists of four components: i) \textbf{ Outlier Embedder (OE)}, ii) \textbf{ Cross-Temporal Scale Transformer (CTST)} and iii) \textbf{Detector}. The weakly-supervised module takes two batches of inputs $F_O$, and $F_M$ for binary classification of typical and atypical videos. 
\\
\subsubsection{Outlier Embedder OE}
To create pseudo-temporal position embeddings that are aware of atypical (autistic behaviors in this case) in untrimmed videos, it is crucial to understand the representations at the typical segment level. This way, any temporal segment that significantly deviates from the established typical patterns is identified as an outlier, or an ASD. In such situations, it makes sense to learn the spatio-temporal cues of videos that belong to a one-class (i.e., typical) distribution. The outlier embedder focuses on understanding the temporal patterns rather than visual signals in non-autistic videos. 
\\
\subsubsection{Cross-Temporal Scale Transformer (CTST)}
The Cross Temporal Scale Transformer (CTST) aims to learn distinct representations for atypical behaviors of varying lengths in relation to their typical counterparts. Given that short and long atypical behaviors are defined by separate cues (i.e., sharp and progressive spatio-temporal cues, respectively), it is advantageous to encode temporal relationships at multiple semantic levels (i.e., temporal scale). The CTST employs a multi-level architecture based on a temporal feature pyramid to accommodate both long- and short-length ASD. The lower levels of the CTST capture the fine-grained, sharp temporal changes associated with short ASD markers, while the higher levels compile the contextual temporal progression of long ASD markers.
\\
\subsubsection{Detector}
The detector is a Multi-Layer Perceptron (MLP) consisting of three fully-connected layers. It takes in video feature maps of dimension $T \times nm$ and assigns ASD scores to each temporal token. The final layer of the MLP contains a single neuron with a sigmoid activation function, which independently ranks each temporal token. Ultimately, the detector produces a score map $S$ of dimension $T \times 1$, which is utilized for ASD detection.

\subsection{Severity Score Regressor}
The proposed shallow architecture is designed to understand both coarse-fine discriminative markers, using ADOS severity score labels as a basis. This shallow architecture consists of two TCN layers followed by three MLP layers. The module accepts inputs from the visual encoder, represented as $T \times D$, and combines them with feature embeddings of size $T \times 128$ from the trained weakly-supervised module to estimate the severity score. Given that ADOS provides a severity score at the video-level for each child, we max-pool the output to compute the final score as shown in Figure \ref{fig:model}. 
\\

\textbf{Weakly-Supervised Architecture Optimization:} The suggested structure, which includes an Outlier Embedder (OE) and a Cross Temporal Scale Transformer (CTST) with a detector, can be trained together using two separate batches of input video feature maps. The visual encoder, similar to the ones used in references \cite{sultani2018real}, is a pre-trained module that is frozen and is only used for feature extraction. The OE, which only takes the typical video feature maps ($F_O$) during training, is optimized with a reconstruction loss as indicated in Equation~\ref{eq:1}. The CTST with the detector considers both typical and ASD video feature maps $F_M \in \mathbb{R}^{T \times nm}$ to calculate typical ($S_t \in R^T$) and ASD ($S_a \in R^T$) temporal token-wise scores. It optimizes itself with a \textit{self-rectifying loss} proposed by [22], as shown in Equations \ref{eq:2} and \ref{eq:3}.

\begin{equation}
    \mathcal{L}_R(F_O) = ||F_O - F_O^R||^2
    \label{eq:1}
\end{equation}

% \begin{equation}
    \begin{multline}
    \label{eq:2}
    \mathcal{L}_D(S_a, St) = \lambda_1 \max(0, 1 - \sum_{i=1}^{T} (S_a^i)+{\sum_{i=1}^{T} (S_t^i))} \\ + \lambda_2||\text{Err}(\text{Typical}) - \text{Err}(\text{Autistic})||
    \end{multline}
% \end{equation}

% \begin{multline}
\begin{equation}
\label{eq:3}
\text{Err}(X) = 
\begin{cases} 
    \frac{1}{T}\sum_{i=1}^{T}(S_{t}^{i} - Y_{t}^{i})^2, & \text{if } X = \text{Typical}, \\
    \underbrace{\forall i, Y_{t}^{i} = \text{Typical}}_{MSE(S_{t})}, \\

    \frac{1}{T}\sum_{i=1}^{T}(S_{a}^{i} - Y_{a}^{i})^2, & \text{if } X = \text{Autistic}, \\
    \forall i, S_a^i < S_\text{{ref}} \Rightarrow Y_a^i = \text{{Typical}}, \\
    \underbrace{\forall i, S_a^i > S_\text{{ref}} \Rightarrow Y_a^i = \text{{Autistic}}}_{MSE(S_a)}
\end{cases}
\end{equation}

\section{Experimental Details}
This section first details the dataset collected for all experiments, called Autism dataset. Then, it provides the experimental details used.

\subsection{Autism Dataset}
The Autism dataset comprises real-life assessment sessions of children, which were conducted by clinicians at a hospital. These sessions, totaling 132 hours, were recorded in accordance with the ADOS-2 protocol to examine the visual behavior of children based on the severity of their autism. Each child was evaluated for potential autism disorder during various interactive ADOS-2 activities. Untrimmed videos were categorized into nine modules, namely, \textit{anniversary, playing with bubbles, playing with ball, construction, demonstration, describing-image, imitation, joint-game, and puzzle}, as per the ADOS evaluation protocol. Each module corresponds to a specific evaluation criterion. For instance, the module 'playing with ball or bubble' assesses repetitive behaviors, while the 'joint-game' analyzes a child's social skills. These experiments utilize a total of 75 unique hour-long videos of children for the study. The dataset is divided according to the subjects (children) and the severity score of each child assessed by the clinicians, as shown in Table \ref{tab:dataset}. Thus, only one child is present in either train or test set. We split the 75 unique videos into train and test sets in a ratio of 85\% and 15\% respectively, keeping a balanced ratio of severity levels in each set. 

The dataset will be made public in modalities such as skeleton, optical-flow and depth information after receiving approval from the ethical team. 

% \begin{flushleft}
\begin{table}[h]
\begin{center}
\resizebox{\linewidth}{!}{%
\renewcommand{\arraystretch}{1.5} % Adjusts the height of the row
\begin{tabular}{l@{\hspace{0.5cm}}c@{\hspace{0.5cm}}c@{\hspace{0.5cm}}c}
\hline
\begin{tabular}[c]{@{}c@{}}Severity\\ Levels\end{tabular} & \begin{tabular}[c]{@{}c@{}}No. of hour-long\\  Videos\end{tabular} & \begin{tabular}[c]{@{}c@{}}No. of segmented\\  modules/videos \end{tabular} &Train/Test \\ \hline
No-autism & 14      & 35     &   27/8     \\
Weak               & 6      & 19       &   16/3   \\
Moderate           & 20      & 52      &    40/12   \\
High               & 35      & 110      &   87/23    \\ \hline
\end{tabular}
}
\end{center}
\vspace{0.7em}
\caption{Autism dataset analysis based on Severity score. The hour-long videos are subdivided into minutes long ADOS modules.}
\label{tab:dataset}
\end{table}
% \end{flushleft}

\begin{table}[h]
\begin{center}
\resizebox{0.7\linewidth}{!}{%
\renewcommand{\arraystretch}{1.5} % Adjust the value to increase or decrease the space between rows
\begin{tabular}{c@{\hspace{2cm}}c}
\hline
Method           & \begin{tabular}[c]{@{}c@{}}Typical / Autistic\\ frame-level AUC (\%)\end{tabular} \\ \hline
Clip-TSA \cite{joo2023clip}         & 60.01                                                                \\
\textbf{OE-CTST} \cite{majhi2024oe} & \textbf{68.58}                                                       \\ \hline
\end{tabular}
}
\end{center}
\caption{WTAL methods results for typical and atypical behavior classifications. }
\label{tab:wtal}
\end{table}

\begin{table}[h]
\begin{center}
\resizebox{\linewidth}{!}{%
\renewcommand{\arraystretch}{1.5} % Adjust the value to increase or decrease the space between rows
\begin{tabular}{c@{\hspace{0.5cm}}c@{\hspace{0.5cm}}c@{\hspace{0.5cm}}c@{\hspace{0.5cm}}c}
\hline
Method & Backbone  & $\uparrow$ Acc. (\%) & $\downarrow$ MAE  & $\downarrow$ MSE  \\ \hline
MS-TCN \cite{farha2019ms} & DinoV2 & 29.88     & 2.69 & 10.31 \\
PDAN \cite{dai2021pdan}& DinoV2 & 45.18     & 2.33 & 
9.73 \\
\textbf{Ours} & DinoV2 & \textbf{48.69}                                                       & \textbf{2.24}                                                      & \textbf{9.55} \\ \hline
MS-TCN\cite{farha2019ms} & VideoMAE-v2& 31.25     & 2.18 & 9.47 \\
PDAN\cite{dai2021pdan} & VideoMAE-v2 & 48.01     & 2.08 & 
8.14 \\ \hline
\textbf{Ours} & VideoMAE-v2 & \textbf{50.88}                                                       & \textbf{1.77}                                                      & \textbf{7.42} \\ \hline
\end{tabular}
}
\end{center}
\vspace{0.7em}
\caption{Experimental results of Supervised methods on Autism dataset. Accuracy is calculated at frame-level for a classification task.}
\label{tab:spervised}
\end{table}

\subsection{Implementation Details}
Before extracting features, we detect and crop child tracklets from videos across frames using SOTA Track-Anything \cite{yang2023track} and AgeFormer \cite{ali2024p} networks. We consider VideoMAE-v2 \cite{wang2023videomae}, and DinoV2\cite{oquab2023dinov2} for spatio-temporal feature extraction. For each 16-frame snippet, a 1408D feature vector is extracted from the backbone pre-trained on Kinetics dataset~\cite{kay2017kinetics} from VideoMAE-v2-giant, and a $T \times 257 \times 1024$ feature vector from the last hidden layer of DinoV2. We pre-process $T$ frames into $32$ averaged temporal length for dimensionality reduction. We use VideoMAE-v2 features for the final experiments due to its robust spatio-temporal features.  

\begin{figure}[h]
\centering
\includegraphics[width=\linewidth]{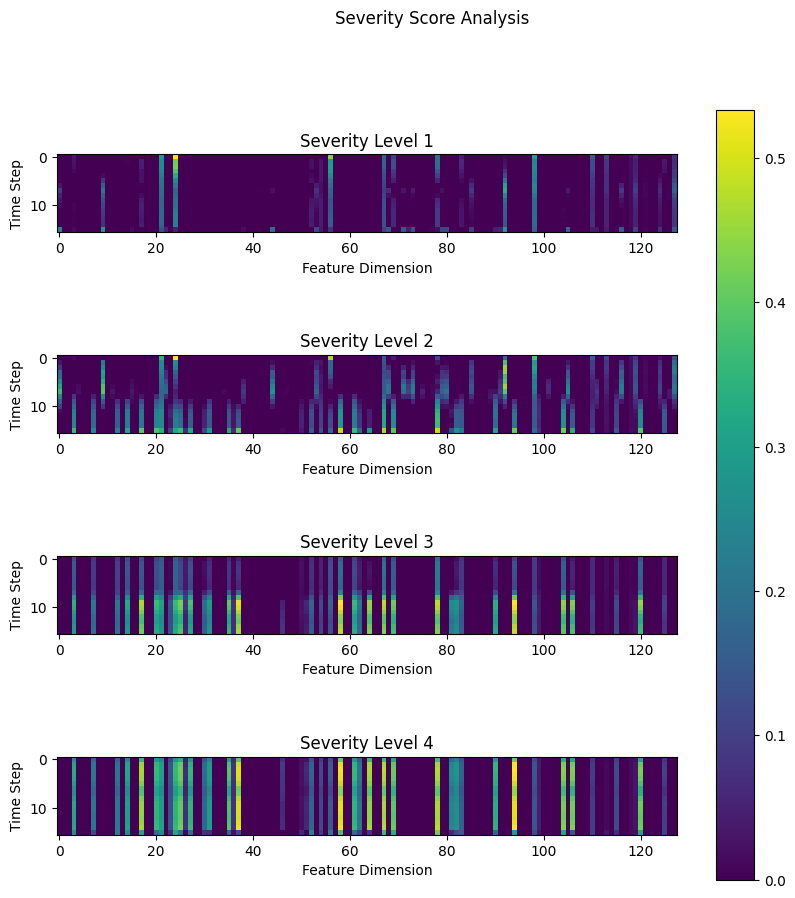}
\caption{Analysis of WTAL $T \times D$ features for 4 randomly selected participants from each level, where $T = 16$ and $D = 128$ (feature vector). The density of the heatmap defines the atypical biomarkers. A higher density on the heatmap corresponds to a higher severity score.}
\label{fig:attn}
\end{figure}
Initially, we adopt the same experimental protocols outlined for OE-CTST in \cite{majhi2024oe} to train a binary classifier, distinguishing between typical and atypical. Training is carried out using the Adam optimizer with a learning rate of $0.001$ over 4000 epochs on our Autism dataset. Upon the completion of OE-CTST training, we freeze the architecture and utilize the 128D embedded features of the Detector for subsequent processing.

Subsequently, we design a shallow TCN-MLP network to enhance learning at the different levels of autism severity. This network is composed of three Temporal Convolution Network (TCN) layers and three MLP layers, which are used to train a score regressor. The TCNs aid in down-sampling the features from the visual encoder, which are then combined with the 128D features of OE-CTST prior to the application of MLP. We employ a supervised training approach for this network, using severity scores as labels over 40 epochs with the Adam optimizer and a learning rate of 0.0001. Furthermore, for the severity score regression we use a ranking loss, specifically Corn Loss \cite{shi2023deep}. We conduct experiments with MSE and MAE for the evaluation of regression scores.

\section{Results and Discussion}
In an hour-long ASD diagnostic session, various atypical biomarkers are observed. These biomarkers represent a range of discriminative patterns, including emotions, repetitive gestures, social interactions, atomic gestures, and unusual movements, among others. Each session is assigned a single severity label. Traditional action recognition methods are not suitable for evaluating or classifying severity scores due to the complexity and diversity of these patterns. As a result, we employ existing Temporal Action Localization (TAL) methods to encode these discriminative markers in long videos. 

\begin{figure}[h]
\centering
\includegraphics[width=0.9\linewidth]{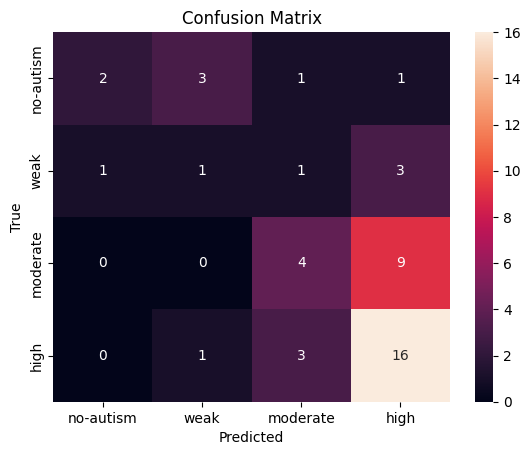}
\caption{Confusion matrix for severity score assessment.}
\label{fig:cm}
\end{figure}

Initially, we conduct experiments with existing supervised TAL methods as depicted in Table \ref{tab:spervised}. PDAN, which are purely TAL methods, did not perform well on the Autism dataset for severity score evaluation. These methods are designed primarily to learn the temporal relations of spatio-temporal features of untrimmed videos. Consequently, applying temporal max-pooling to the last feature embedding layer of these methods did not yield the desired results for severity score computation. Another key factor for the effectiveness of these methods is the availability of densely annotated data, either at the frame-level or segment-level, for each action class in an untrimmed video. As we do not have these annotations, we opt for the Weakly Supervised Temporal Action Localization (WTAL) method. These methods are capable of learning various discriminative biomarkers (both known and unknown) in a weakly supervised setting, thereby enabling the model to discern between typical and atypical behavior patterns. The features derived from the WTAL method are then used to train a regression model for the final score. This proposed approach is proven to be successful, achieving the highest accuracy.

Clip-TSA and OE-CTST, which are state-of-the-art Weakly Supervised Temporal Action Localization (WTAL) methods, are used for anomaly detection in untrimmed videos. We have adapted these methods to learn a binary classification between typical and ASD behavior patterns, as shown in Table \ref{tab:wtal}. OE-CTST outperformed Clip-TSA due to its specialized Outlier Positional Embedding and CTST modules. To illustrate the effectiveness of the OE-CTST network, we visualized the last feature vector of the Detector $T \times 128D$ (T=32) of four randomly selected participants for each severity level, as shown in Figure \ref{fig:attn}. Figure \ref{fig:attn} shows a denser heatmap for videos with a higher severity score, validating the proposed WTAL architecture's suitability for this task. It also demonstrates the amount of biomarkers we identified. For example, for participant having higher severity score, we identify around 50 biomarkers. However, not all biomarkers related to ASD could be identified and is left for future work. Based on these identified biomarkers and features from the visual encoder we train a supervised network on top of the WTAL for the final regression of the severity score, as shown in Table \ref{tab:spervised}.

The confusion matrix computed on test-set depicted in Figure \ref{fig:cm} offers a comprehensive insight into the evaluations of severity assessment. The model exhibits superior performance for the \textit{high} and \textit{moderate} classes in comparison to the \textit{no-autism} and \textit{weak} classes. This performance can be attributed to the higher correlations between these classes, as illustrated in Figure \ref{fig:attn}. Furthermore, the one outlier confusion between the \textit{high} class and the \textit{no-autism} classes is because the child is not autistic but hyperactive. We present and deliberate on this particular case with the clinician to ascertain whether it is an error in the analysis or if the child is genuinely enthusiastic about playing with bubbles and does not have autism. The clinicians confirmed that this child is merely extroverted and hyperactive. However, such scenarios can lead the model to mistakenly identify a higher autism case for hyperactive children. As a result, we plan to introduce an additional class for hyperactive cases in the future to prevent such inaccuracies.
% After visualizing the video (playing with ball and bubbles) along with the clinici, we notice the child is excited to play with bubbles.
% The confusion matrix in Figure \ref{fig:cm} provides in-depth understanding of severity assessment evaluations. The dataset is highly imbalanced which reflects in the confusion matrix. Despite that, our model can predict some of "high", and "moderate" scores accurately. Additionally, the confusion between these two classes 
\section{Conclusion}
Capturing autistic biomarkers without dense annotations is a challenging task, particularly in long untrimmed videos. The wide array of biomarkers, such as facial expressions, uncontrolled movements, repetitive behaviors, and eye-gaze, present in a long video with a single severity label, complicates accurate detection by the model. Additionally, the complexity is further increased by human errors and the subjectivity of the severity score. In this study, we strive to learn these discriminative markers in a weakly-supervised manner for atypical behaviors, which are then divided into four distinct severity levels for ASD evaluations. Despite these challenges, our proposed method achieves the highest accuracy compared to the baseline results.
Our method, which is based on WTAL, offers numerous advantages. It provides clinicians with a tool to validate these biomarkers, enabling them to make more objective decisions. In addition, it aids clinicians to perform a comprehensive diagnosis by considering all discriminative biomarkers, known and unknown. This can be highly beneficial.

{\small
\bibliographystyle{IEEEtran}
\bibliography{egbib}
}

\end{document}